\documentclass{article}

\usepackage[preprint]{neurips_2026}

\usepackage[utf8]{inputenc}
\usepackage[T1]{fontenc}
\usepackage{hyperref}
\usepackage{url}
\usepackage{booktabs}
\usepackage{amsfonts}
\usepackage{amsmath}
\usepackage{amssymb}
\usepackage{nicefrac}

\usepackage{microtype}
\usepackage{xcolor}
\usepackage{graphicx}
\usepackage{multirow}
\usepackage{algorithm}
\usepackage{algpseudocode}
\usepackage{subcaption}
\usepackage{float}

\title{TheProfessor: Multi-Teacher Unsupervised Prompt Distillation\\for Vision-Language Models}

\author{%
  Ahmad Algadhi \\  s202258940@kfupm.edu.sa \And
  Ahmed M. Alzuhair \\ s202278540@kfupm.edu.sa \And
  Omar Alkhulaif \\ s202167450@kfupm.edu.sa \And
  Muzammil Behzad \\ muzammil.behzad@kfupm.edu.sa \\
  Department of Information and Computer Science \\
  King Fahd University of Petroleum and Minerals
}

\begin{document}

\maketitle

\begin{abstract}
Prompt distillation compresses large vision-language models (VLMs) such as CLIP into lightweight student models by matching teacher predictions on unlabeled domain images. PromptKD (CVPR 2024) established this paradigm with a single PromptSRC-finetuned ViT-L/14 teacher and a ViT-B/16 student. We propose \textbf{TheProfessor}, a multi-teacher extension that distills from a fixed two-teacher ensemble: a domain-finetuned PromptSRC ViT-L/14 teacher and a zero-shot EVA-CLIP-L/14 teacher whose logits are pre-computed per dataset. We evaluate single-teacher PromptKD, equal-probability ensembling, and confidence-weighted ensembling on four base-to-novel datasets: Caltech-101, DTD, UCF101, and EuroSAT. In a 12-run single-seed sweep, confidence-weighted ensembling improves average HM from 87.52 to 89.28 (+1.77 points), while equal averaging improves average HM to 88.88 (+1.37 points). Gains are dataset dependent: they are negligible on Caltech-101 (+0.16 HM for confidence weighting), modest on UCF101 (+0.62), and largest on domain-shifted EuroSAT (+5.78). These results update our earlier Caltech-only analysis and show that multi-teacher prompt distillation is most useful when the second teacher contributes complementary supervision under domain shift.
\end{abstract}

\section{Introduction}

Large-scale vision-language models (VLMs) such as CLIP~\cite{radford2021learning} have transformed visual recognition by learning transferable image-text representations from web-scale data. However, deploying large CLIP variants such as ViT-L/14 can be expensive. PromptKD~\cite{li2024promptkd} addresses this by distilling a large PromptSRC-finetuned CLIP teacher into a lighter ViT-B/16 student using unlabeled domain images and a KL-divergence loss. The teacher provides soft class distributions, while the student learns prompt parameters and a projection layer; at inference time, the expensive teacher is removed.

A limitation of PromptKD is its reliance on a \emph{single} teacher. A single teacher encodes one inductive bias over a domain: its pretraining data, architecture, prompt tuning objective, and calibration all affect the soft labels used for distillation. Multi-teacher distillation has long been used to combine complementary teacher signals~\cite{you2017learning,liu2020adaptive,yuan2021revisiting}. This motivates the central question of our work: \emph{can prompt distillation for VLMs benefit from a heterogeneous teacher ensemble?}

We introduce \textbf{TheProfessor}, a multi-teacher extension of PromptKD that combines two CLIP-family teachers:
\begin{itemize}
    \item \textbf{Teacher 1 (T1):} a PromptSRC-finetuned CLIP ViT-L/14 teacher, matching the original PromptKD setup.
    \item \textbf{Teacher 2 (T2):} EVA-CLIP-L/14~\cite{sun2023eva}, a zero-shot teacher trained with a different recipe and data mixture. T2 is not finetuned on the target domain; its logits are pre-computed and cached for every unlabeled training image.
\end{itemize}

We evaluate two ensemble strategies: \textbf{prediction averaging}, which gives both teachers equal weight, and \textbf{confidence-weighted averaging}, which gives more influence to the teacher with the higher maximum softmax confidence for each image. Unlike an online ensemble that evaluates both large teachers at every training step, our implementation caches T2 logits once per dataset and retrieves them by image path during student training.

Our updated experiments span four datasets: Caltech-101, DTD, UCF101, and EuroSAT. The results show that multi-teacher gains depend strongly on dataset/domain characteristics. On Caltech-101, the method is essentially tied with the single-teacher baseline; on DTD and UCF101 it gives modest improvements; on EuroSAT, confidence-weighted ensembling improves HM by +5.78 points. Averaged across the four datasets, confidence weighting improves HM by +1.77 points and equal averaging by +1.37 points.

Our contributions are:
\begin{itemize}
    \item We propose TheProfessor, a cache-efficient multi-teacher extension of PromptKD for CLIP-style VLM prompt distillation.
    \item We implement and evaluate two ensemble modes, equal averaging and confidence weighting, using a fixed PromptSRC teacher and a cached EVA-CLIP teacher.
    \item We report an updated four-dataset single-seed sweep showing that multi-teacher distillation is most beneficial on domain-shifted datasets, especially EuroSAT.
    \item We refine the earlier Teacher Agreement Ceiling hypothesis: high teacher agreement can explain negligible gains on easy datasets, while lower complementarity barriers on shifted domains can yield substantial improvements.
\end{itemize}
\section{Related Work}

\paragraph{Prompt Learning for Vision-Language Models.}
Prompt learning adapts large pre-trained VLMs to downstream tasks by optimizing a small set of continuous prompt tokens rather than the full model weights~\cite{lester2021power,li2021prefix}. CoOp~\cite{zhou2022learning} pioneered learnable text prompts for CLIP; CoCoOp~\cite{zhou2022conditional} conditioned them on visual input; MaPLe~\cite{khattak2023maple} extended prompt tuning to both modalities simultaneously. PromptSRC~\cite{khattak2023self} introduced self-regularization to prevent forgetting of zero-shot generalization. These methods rely on labeled few-shot data and cannot exploit the large pools of unlabeled domain images available in practice.

\paragraph{Knowledge Distillation.}
Knowledge distillation~\cite{hinton2015distilling} trains a lightweight student under the supervision of a larger teacher's soft predictions. Feature-based~\cite{romero2014fitnets,chen2022knowledge}, logit-based~\cite{zhao2022decoupled}, and relation-based~\cite{park2019relational} variants have all been proposed. CLIP-KD~\cite{yang2023clip} and TinyCLIP~\cite{wu2023tinyclip} applied distillation specifically to CLIP models. PromptKD~\cite{li2024promptkd} uniquely combines prompt learning with unsupervised distillation, training only the student's visual prompts rather than full-model fine-tuning. Our work extends this line by introducing multi-teacher supervision.

\paragraph{Multi-Teacher Knowledge Distillation.}
Ensembling multiple teacher predictions has been shown to improve student training across image classification~\cite{you2017learning,fukuda2017efficient}, natural language processing~\cite{clark2019bam}, and model compression~\cite{liu2020adaptive}. Yuan et al.~\cite{yuan2021revisiting} showed that label smoothing from weaker teachers can act as a form of knowledge regularization. However, a consistent finding in the literature is that teacher diversity --- measured by disagreement between teachers on held-out data --- is the key driver of multi-teacher gain~\cite{malinin2019ensemble}. To date, no work has examined multi-teacher distillation in the VLM prompt learning context.

\paragraph{EVA-CLIP.}
EVA-CLIP~\cite{sun2023eva} is a family of CLIP-style models trained on a mixture of LAION-2B and COYO-700M, using an EVA~\cite{fang2023eva} visual backbone that incorporates masked image modeling pre-training. EVA-CLIP models exhibit strong zero-shot generalization across diverse benchmarks, making them appealing as complementary teachers to OpenAI CLIP.

\section{Method}

\subsection{Background: PromptKD}

PromptKD~\cite{li2024promptkd} proceeds in two stages. In Stage I, a large CLIP teacher $f_T = (f_T^I, f_T^{\text{txt}})$ is fine-tuned on domain few-shot labeled data using PromptSRC, producing well-trained text features $\mathbf{W} = [w_1, \dots, w_N] \in \mathbb{R}^{N \times d}$ for all $N$ classes. These features are pre-stored and frozen for Stage II. In Stage II, a lightweight student image encoder $f_S^I$ with learnable visual prompts is trained on unlabeled domain images by minimizing the KL divergence between teacher and student logits:
\begin{equation}
\mathcal{L}_{\text{stu}} = \tau^2 \cdot \text{KL}\!\left(\sigma(q^t/\tau),\; \sigma(q^s/\tau)\right),
\label{eq:kd}
\end{equation}
where $q^t = u^t \mathbf{W}^\top$, $q^s = P(u^s) \mathbf{W}^\top$ are teacher and student logits, $P(\cdot)$ is a feature projector handling dimension mismatch, and $\tau$ is the temperature.

\subsection{TheProfessor: Multi-Teacher Extension}

We extend Stage II to incorporate a second teacher T2. Let $q^{t_1}$ and $q^{t_2}$ be the logits produced by T1 and T2 on image $x$, and let $p^{t_i}=\sigma(q^{t_i}/\tau)$ denote the softened distribution from teacher $i$. We define the \textbf{ensemble teacher distribution} as:

\paragraph{Average Ensemble (Avg).}
\begin{equation}
    p^{\text{ens}}_{\text{avg}} = \frac{1}{2}\left(p^{t_1} + p^{t_2}\right),
    \label{eq:avg}
\end{equation}
The KL loss is then $\mathcal{L}_{\text{stu}} = \tau^2 \cdot \text{KL}(p^{\text{ens}}_{\text{avg}}, \sigma(q^s/\tau))$.

\paragraph{Confidence-Weighted Ensemble (Conf).}
The implementation weights teachers by their maximum softmax confidence on each image. Let $c_i = \max_c p^{t_i}_c$. The weights and ensemble distribution are:
\begin{equation}
    \alpha_i = \frac{c_i}{c_1 + c_2 + 10^{-8}},
\end{equation}
\begin{equation}
    p^{\text{ens}}_{\text{conf}} = \alpha_1 p^{t_1} + \alpha_2 p^{t_2}.
\end{equation}

\paragraph{Efficient T2 Integration via Caching.}
T2 (EVA-CLIP-L/14) is a large model with 768-d embeddings. To avoid running T2 at every training step, we pre-compute and cache its predictions on all training images in a single forward pass before Stage II begins. For each training image $x_i$, we store the T2 logits used to form $p^{t_2}_i$. During training, T2 logits are retrieved from the cache by image path at zero additional GPU cost. This is analogous to PromptKD's text feature caching trick applied to the visual domain.

\begin{figure}[H]
\centering
\includegraphics[width=\linewidth]{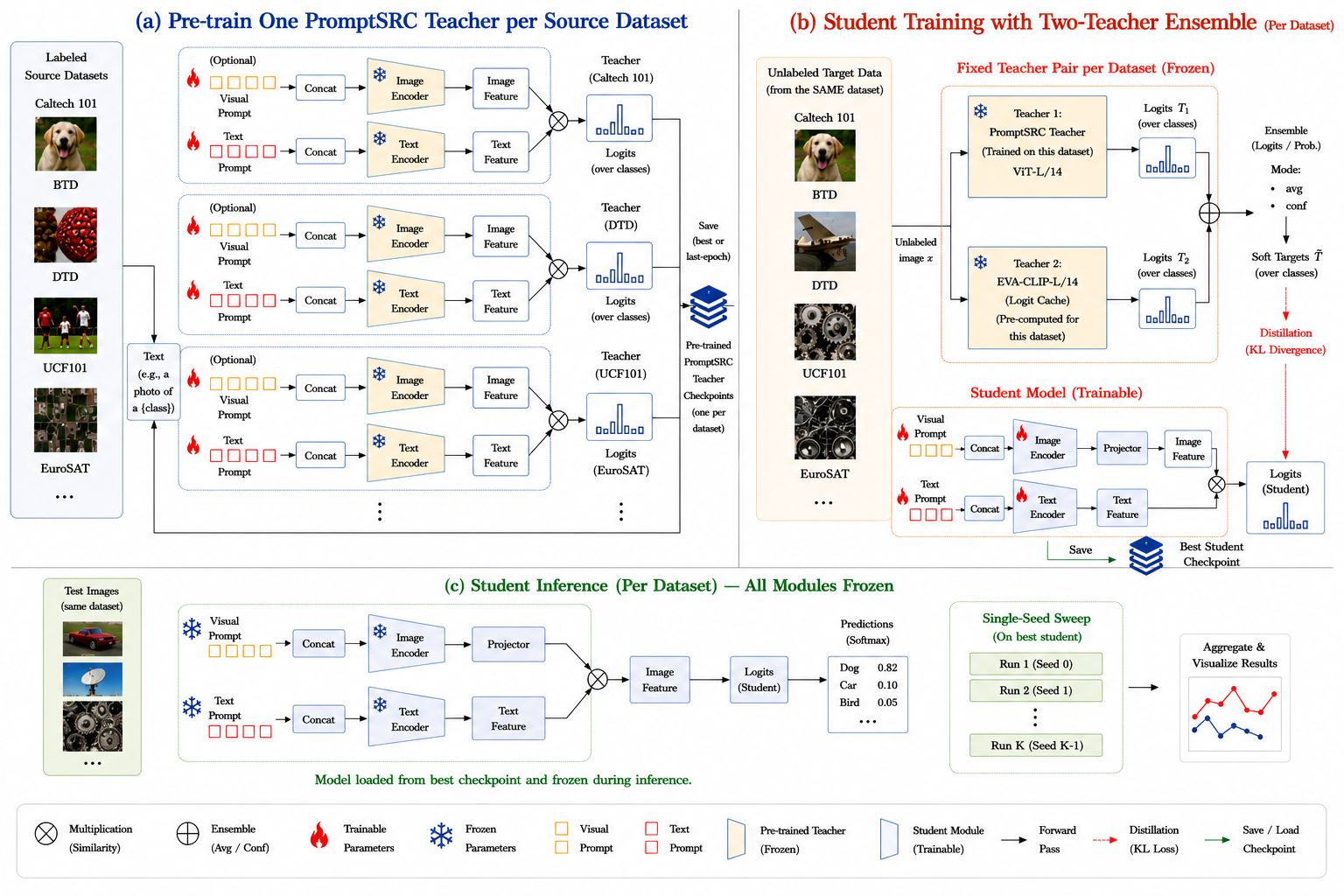}
\caption{\textbf{TheProfessor architecture.} Stage I pre-trains one PromptSRC teacher per source dataset using labeled images. Stage II trains a ViT-B/16 student on unlabeled images from the same dataset using a frozen PromptSRC teacher and cached EVA-CLIP-L/14 logits. The ensemble target is formed by equal averaging or confidence weighting, and the student is optimized with KL divergence. At inference time, only the trained student is used.}
\label{fig:architecture}
\end{figure}

\paragraph{Architecture Overview.}
Figure~\ref{fig:architecture} illustrates the full TheProfessor pipeline. Stage I is unchanged from PromptKD: T1 is fine-tuned with PromptSRC on labeled data, and its text features are stored. Stage II adds a T2 logit cache and a multi-teacher KL loss. At inference, only the student ViT-B/16 and the stored T1 text features are used --- identical to single-teacher PromptKD.

\subsection{Pseudocode}

Algorithm~\ref{alg:theprofessor} summarizes the TheProfessor training procedure.

\begin{algorithm}[H]
\caption{TheProfessor: Multi-Teacher Prompt Distillation}
\label{alg:theprofessor}
\begin{algorithmic}[1]
\Require Unlabeled images $\{x_i\}$, T1 (PromptSRC ViT-L/14), T2 (EVA-CLIP-L/14), student $f_S^I$
\State $\mathbf{W} \gets f_{T1}^{\text{txt}}(\text{classnames})$ \Comment{Pre-store T1 text features}
\State $\{q^{t_2}_i\} \gets \text{CacheT2}(\{x_i\}, T2)$ \Comment{Pre-cache T2 logits}
\For{each mini-batch $(x_i, \text{path}_i)$}
    \State $q^{t_1}_i \gets f_{T1}^I(x_i) \cdot \mathbf{W}^\top$ \Comment{T1 logits (live)}
    \State $q^{t_2}_i \gets \text{lookup}(\text{path}_i)$ \Comment{T2 logits (cached)}
    \State $p^{\text{ens}}_i \gets \text{Ensemble}(q^{t_1}_i, q^{t_2}_i)$ \Comment{Avg or Conf}
    \State $q^s_i \gets P(f_S^I(x_i)) \cdot \mathbf{W}^\top$ \Comment{Student logits}
    \State $\mathcal{L} \gets \tau^2 \cdot \text{KL}(p^{\text{ens}}_i,\; \sigma(q^s_i/\tau))$
    \State Update student prompts and projector via $\nabla \mathcal{L}$
\EndFor
\State \Return student $f_S^I$, $\mathbf{W}$
\end{algorithmic}
\end{algorithm}

\section{Experiments}

\subsection{Experimental Setup}

\paragraph{Datasets.}
We evaluate on four PromptKD base-to-novel benchmarks: \textbf{Caltech-101}~\cite{fei2004learning}, \textbf{Describable Textures Dataset (DTD)}, \textbf{UCF101}, and \textbf{EuroSAT}. Each dataset follows the PromptKD/Dassl base-to-novel protocol: the PromptSRC teacher is trained on base classes using 16-shot labels, while the student is trained on unlabeled target-domain images and evaluated on base and novel splits. The T2 cache contains all training images used by the student: 4,128 images for Caltech-101, 2,820 for DTD, 7,639 for UCF101, and 13,500 for EuroSAT.

\paragraph{Models.}
\begin{itemize}
    \item \textbf{Teacher 1 (T1):} CLIP ViT-L/14 with PromptSRC-learned prompts. The Caltech-101 teacher is the released PromptKD teacher; DTD, UCF101, and EuroSAT teachers are trained with PromptSRC using the same ViT-L/14 configuration.
    \item \textbf{Teacher 2 (T2):} EVA-CLIP-L/14 (\texttt{EVA02-L-14/merged2b\_s4b\_b131k}), used zero-shot. We pre-compute T2 logits for all student-training images and store them in a per-dataset cache.
    \item \textbf{Student:} CLIP ViT-B/16 with learnable visual and text prompts (4 context tokens, prompt depth 9), matching the PromptKD student configuration.
\end{itemize}

\paragraph{Implementation Details.}
We modify the PromptKD codebase with three configuration flags: \texttt{USE\_T2}, \texttt{T2\_CACHE}, and \texttt{ENSEMBLE\_MODE}. Training uses 20 epochs, batch size 8, temperature $\tau=1.0$, and the PromptKD ViT-B/16 configuration. Dataset-specific KD weights follow the sweep script: 1000 for Caltech-101 and UCF101, 200 for DTD, and 1000 for EuroSAT. The full sweep contains 12 runs: four datasets $\times$ three methods (single teacher, multi-teacher average, multi-teacher confidence). All numbers are reported for seed 1.

\paragraph{Evaluation Metrics.}
We report Base accuracy, Novel accuracy, and their harmonic mean (HM):
\begin{equation}
    \mathrm{HM} = \frac{2 \cdot \mathrm{Base} \cdot \mathrm{Novel}}{\mathrm{Base} + \mathrm{Novel}}.
\end{equation}

\subsection{Teacher and Cache Preparation}

Before running the sweep, we verify that all T1 teachers and all T2 caches are available. The Caltech PromptSRC sanity check reaches 98.5\% base validation accuracy, confirming that the PromptSRC recipe is healthy. The missing DTD, UCF101, and EuroSAT teachers are then trained successfully and copied into the expected PromptKD teacher directories. For T2, we use a custom pre-computation dataloader with \texttt{drop\_last=False}; this avoids silently missing the final partial batch and ensures that the cache size exactly matches \texttt{train\_x} for each dataset.

\begin{table}[h]
\centering
\caption{\textbf{T2 cache coverage.} The cached EVA-CLIP logits cover every unlabeled training image used by the student.}
\label{tab:cache}
\begin{tabular}{lrrr}
\toprule
\textbf{Dataset} & \textbf{Classes} & \textbf{Train images} & \textbf{Cached logits} \\
\midrule
Caltech-101 & 100 & 4,128 & 4,128 \\
DTD & 47 & 2,820 & 2,820 \\
UCF101 & 101 & 7,639 & 7,639 \\
EuroSAT & 10 & 13,500 & 13,500 \\
\bottomrule
\end{tabular}
\end{table}

\subsection{Main Results}

Table~\ref{tab:main} reports the updated four-dataset single-seed sweep. Multi-teacher ensembling improves the harmonic mean on three of four datasets under confidence weighting and on three of four datasets under equal averaging. The largest gains occur on EuroSAT, where the confidence-weighted ensemble improves HM from 89.57 to 95.35 (+5.78).

\begin{table}[t]
\centering
\caption{\textbf{Base-to-novel results for the single-seed sweep.} ``Single'' is PromptKD with only the PromptSRC T1 teacher. ``Avg'' averages the T1 and T2 probability distributions. ``Conf'' weights teachers by confidence. $\Delta$HM is relative to the single-teacher baseline.}
\label{tab:main}
\resizebox{\linewidth}{!}{%
\begin{tabular}{llrrrr}
\toprule
\textbf{Dataset} & \textbf{Method} & \textbf{Base (\%)} & \textbf{Novel (\%)} & \textbf{HM (\%)} & \textbf{$\Delta$HM} \\
\midrule
\multirow{3}{*}{Caltech-101}
 & Single & 98.70 & 96.50 & 97.59 & -- \\
 & Avg & 98.30 & 96.70 & 97.49 & -0.09 \\
 & Conf & 98.60 & 96.90 & 97.74 & +0.16 \\
\midrule
\multirow{3}{*}{DTD}
 & Single & 85.30 & 72.60 & 78.44 & -- \\
 & Avg & 85.20 & 75.70 & 80.17 & +1.73 \\
 & Conf & 85.00 & 73.70 & 78.95 & +0.51 \\
\midrule
\multirow{3}{*}{UCF101}
 & Single & 89.10 & 80.30 & 84.47 & -- \\
 & Avg & 86.80 & 82.40 & 84.54 & +0.07 \\
 & Conf & 87.40 & 82.90 & 85.09 & +0.62 \\
\midrule
\multirow{3}{*}{EuroSAT}
 & Single & 97.00 & 83.20 & 89.57 & -- \\
 & Avg & 96.00 & 90.80 & 93.33 & +3.76 \\
 & Conf & 95.30 & 95.40 & 95.35 & +5.78 \\
\bottomrule
\end{tabular}%
}
\end{table}

\subsection{Aggregate Results}

Table~\ref{tab:aggregate} summarizes mean performance across the four datasets. Equal averaging improves mean novel accuracy by +3.25 points and mean HM by +1.37 points. Confidence weighting improves mean novel accuracy by +4.07 points and mean HM by +1.77 points. The base accuracy drop is expected: the second teacher is zero-shot and not specialized to the base classes, so the ensemble trades a small amount of base performance for substantially better novel performance.

\begin{table}[h]
\centering
\caption{\textbf{Mean performance across four datasets.} Confidence weighting gives the best average HM.}
\label{tab:aggregate}
\begin{tabular}{lrrrr}
\toprule
\textbf{Method} & \textbf{Base (\%)} & \textbf{Novel (\%)} & \textbf{HM (\%)} & \textbf{Mean $\Delta$HM} \\
\midrule
Single teacher & 92.52 & 83.15 & 87.52 & -- \\
Multi-teacher Avg & 91.58 & 86.40 & 88.88 & +1.37 \\
Multi-teacher Conf & 91.58 & 87.22 & 89.28 & +1.77 \\
\bottomrule
\end{tabular}
\end{table}

\subsection{Key Observations}

\paragraph{Caltech-101 remains near the teacher-agreement ceiling.}
The Caltech-101 results are effectively tied: Avg is -0.09 HM and Conf is +0.16 HM relative to the single-teacher baseline. This supports the original observation that easy object-recognition datasets leave little room for a second large VLM teacher to add information.

\paragraph{Domain-shifted datasets benefit most.}
EuroSAT shows the strongest gain. The confidence-weighted ensemble raises novel accuracy from 83.20\% to 95.40\%, producing a +5.78 HM improvement. This suggests that the zero-shot EVA-CLIP teacher contributes complementary supervision on satellite imagery, where the PromptSRC teacher and student benefit from broader visual priors.

\paragraph{Confidence weighting is more robust than equal averaging on average.}
Equal averaging performs best on DTD, but confidence weighting is best on Caltech-101, UCF101, and EuroSAT, and it achieves the best overall mean HM. This supports using per-image confidence as a simple, low-cost routing signal when the relative quality of teachers varies by dataset and image.

\section{Analysis: Teacher Diversity and Domain Shift}

\subsection{Revisiting the Teacher Agreement Ceiling}

Let $p^{t_1}$ and $p^{t_2}$ be the probability distributions from T1 and T2, and let $p^{\mathrm{ens}} = \frac{1}{2}(p^{t_1}+p^{t_2})$. The difference between single-teacher and ensemble supervision is governed by the Jensen-Shannon divergence between the two teachers:
\begin{equation}
    \mathrm{JSD}(p^{t_1}, p^{t_2}) = \frac{1}{2}\mathrm{KL}(p^{t_1}\|p^{\mathrm{ens}}) + \frac{1}{2}\mathrm{KL}(p^{t_2}\|p^{\mathrm{ens}}).
    \label{eq:jsd}
\end{equation}
If $\mathrm{JSD}(p^{t_1},p^{t_2}) \approx 0$, the ensemble distribution is nearly identical to either teacher and the student receives little additional information. This is the Teacher Agreement Ceiling: when two strong teachers make nearly the same predictions with similar confidence, ensembling is unlikely to improve student accuracy.

The updated sweep refines this hypothesis. Caltech-101 is still a ceiling case, but EuroSAT is not. The observed pattern suggests that teacher diversity is not a global property of a teacher pair; it is dataset dependent. The same T1+T2 pair can be redundant on canonical object datasets and complementary on shifted domains.

\subsection{Why EuroSAT Improves Most}

EuroSAT differs substantially from the natural-image distributions used in common object recognition benchmarks. Satellite images contain overhead viewpoints, land-use textures, and scene-level patterns. In this setting, the zero-shot EVA-CLIP teacher appears to provide useful probability mass for novel classes, while confidence weighting prevents it from dominating when T1 is more reliable. The result is a large gain in novel accuracy (+12.20 points for Conf) with a smaller base-accuracy tradeoff (-1.70 points), yielding the best HM improvement in the sweep.

\subsection{Base-Novel Tradeoff}

Multi-teacher ensembling consistently improves novel accuracy more than base accuracy. This is expected because T1 is explicitly trained on base classes via PromptSRC, while T2 remains zero-shot. The ensemble partially smooths the base-specialized T1 targets with a more generalist teacher. This can slightly reduce base accuracy but improve transfer to novel classes. Since HM penalizes imbalance, the method is most successful when the novel improvement outweighs the base drop, as in EuroSAT.

\subsection{Implications for Teacher Selection}

The results suggest a practical protocol for multi-teacher prompt distillation:
\begin{enumerate}
    \item Evaluate candidate teachers on a small held-out calibration split or unlabeled proxy set.
    \item Measure top-1 agreement, confidence disagreement, and/or JSD between teachers.
    \item Prefer teacher pairs that are accurate but not redundant, especially for domain-shifted or fine-grained datasets.
    \item Use confidence weighting when teacher reliability varies across images.
\end{enumerate}

\section{Discussion}

\paragraph{When does multi-teacher distillation help?}
The four-dataset sweep shows that multi-teacher distillation helps when the second teacher contributes complementary novel-class supervision. It does not automatically help just because another large model is added. Caltech-101 demonstrates the low-gain regime, while EuroSAT demonstrates the high-gain regime.

\paragraph{Average versus confidence-weighted ensembling.}
Equal averaging is simple and works well when both teachers have similar reliability, as seen on DTD. Confidence weighting is more adaptive: it allows the ensemble to shift toward the teacher that appears more certain for a given image. This makes it the best method on average in our sweep.

\paragraph{Efficient deployment.}
TheProfessor preserves PromptKD's inference-time efficiency. T2 is only used to generate cached training logits; it is never needed at inference. The final model is still the lightweight student, so multi-teacher training improves supervision without increasing deployment cost.

\paragraph{Limitations.}
Our updated experiments are still single-seed. Although the trends are clear, multi-seed averaging is needed to quantify variance, especially for small gains on Caltech-101 and UCF101. We also test only one T2 model and two simple ensemble strategies. Future work should evaluate learned weighting, teacher calibration, additional specialist teachers, and all 11 PromptKD datasets.

\section{Conclusion}

We introduced TheProfessor, a cache-efficient multi-teacher extension of PromptKD for CLIP-based vision-language prompt distillation. The updated results from the MultiTeacher PromptKD single-seed sweep show that adding a zero-shot EVA-CLIP-L/14 teacher can improve a PromptKD student when the dataset benefits from complementary supervision. Across Caltech-101, DTD, UCF101, and EuroSAT, equal averaging improves mean HM by +1.37 points and confidence weighting improves mean HM by +1.77 points. The largest gain occurs on EuroSAT, where confidence weighting improves HM by +5.78 points. These findings revise the earlier Caltech-only conclusion: teacher agreement can cap gains on easy datasets, but domain shift can unlock substantial multi-teacher benefits without increasing inference cost.
\begin{ack}
This work was supported by ICS 483 Computer Vision at King Fahd University of Petroleum and Minerals. The authors thank Dr. Zheng Li and colleagues for open-sourcing the PromptKD codebase and pre-trained teacher checkpoints. Experiments were conducted on an NVIDIA H100 80GB GPU.
\end{ack}

\appendix

\section{Additional Implementation Details}
\label{app:impl}

\paragraph{PromptKD configuration.}
We use the official PromptKD ViT-B/16 configuration \texttt{vit\_b16\_c2\_ep20\_batch8\_4+4ctx.yaml}: 4 vision context tokens, 4 text context tokens, prompt depth 9, 20 training epochs, SGD with cosine annealing, and batch size 8.

\paragraph{Code patches.}
Running the code on the experiment environment required PyTorch 2.x compatibility patches, including loading checkpoints with \texttt{weights\_only=False} and removing deprecated scheduler arguments. The multi-teacher extension adds three config flags: \texttt{USE\_T2}, \texttt{T2\_CACHE}, and \texttt{ENSEMBLE\_MODE}. In \texttt{trainers/promptkd.py}, the student trainer optionally loads a T2 cache and blends T1 and T2 probability distributions before computing the KL loss.

\paragraph{Teacher training.}
The released Caltech-101 PromptKD teacher is used for Caltech-101. PromptSRC ViT-L/14 teachers are trained for DTD, UCF101, and EuroSAT. Because PromptSRC may not always write \texttt{model-best.pth.tar}, the sweep uses the final epoch checkpoint as a fallback when needed. A Caltech sanity check reaches 98.5\% base validation accuracy before training the missing teachers.

\paragraph{T2 logit caching.}
The T2 caching script uses EVA-CLIP-L/14 to encode all student-training images with deterministic transforms. It uses \texttt{drop\_last=False} and asserts that the number of cached image paths equals the number of training images. The resulting cache sizes are 4,128 (Caltech-101), 2,820 (DTD), 7,639 (UCF101), and 13,500 (EuroSAT).

\section{Per-Dataset Result Summary}
\label{app:results}

\begin{table}[h]
\centering
\caption{Best method by dataset in the single-seed sweep.}
\begin{tabular}{lccc}
\toprule
\textbf{Dataset} & \textbf{Best method} & \textbf{Best HM (\%)} & \textbf{Gain over single} \\
\midrule
Caltech-101 & Confidence-weighted & 97.74 & +0.16 \\
DTD & Average & 80.17 & +1.73 \\
UCF101 & Confidence-weighted & 85.09 & +0.62 \\
EuroSAT & Confidence-weighted & 95.35 & +5.78 \\
\bottomrule
\end{tabular}
\end{table}

\section{Notes on Single-Seed Interpretation}
\label{app:seed}

The current sweep is intended as a reproducible, end-to-end validation of the multi-teacher implementation. Since all values are single-seed results, small differences such as Caltech-101 Avg (-0.09 HM) or UCF101 Avg (+0.07 HM) should be interpreted cautiously. Larger effects, especially EuroSAT Conf (+5.78 HM), are more likely to reflect a meaningful benefit from multi-teacher supervision.

\end{document}